%% file: main.tex
\documentclass[letterpaper, 10 pt, conference]{ieeeconf}  %

\IEEEoverridecommandlockouts                              %

\usepackage{hyperref}
\hypersetup{
    colorlinks=true,
    linkcolor=blue,
    filecolor=magenta,      
    urlcolor=cyan,
    pdftitle={Overleaf Example},
    pdfpagemode=FullScreen,
    }
\usepackage{gensymb}   
\usepackage{times}
\usepackage{epsfig}
\usepackage{graphicx}
\usepackage{amsmath,amssymb,amsfonts}
\usepackage{booktabs}
\usepackage[noend]{algpseudocode}
\usepackage{balance}
\usepackage{xcolor}
\usepackage{url}
\usepackage{authblk}
\usepackage{multirow}
\newcommand{\eg}{\emph{e.g.},}
\newcommand{\ie}{\emph{i.e.},}

\def\secref#1{Section~\ref{#1}}
\def\figref#1{Figure~\ref{#1}}
\def\tabref#1{Table~\ref{#1}}
\def\eqref#1{Equation~(\ref{#1})}

\usepackage{arydshln}
\usepackage{cite}
\usepackage{nicematrix}
\usepackage{subcaption}
\usepackage{graphbox}
\usepackage{soul}
\usepackage{svg}
\usepackage{color, colortbl}
\usepackage{tabularx}

\usepackage[linesnumbered,ruled]{algorithm2e}
\usepackage[flushleft]{threeparttable}
\usepackage{lipsum}
\usepackage{amssymb}%
\usepackage{pifont}%
\newcommand{\cmark}{\ding{51}}%
\newcommand\numberthis[1][]{%
    \refstepcounter{equation}%
    \ifx#1\empty\else\label{eq:#1}\fi%
    \tag{\theequation}%
}

\def\red{\textcolor{black}}

\makeatletter
\def\BState{\State\hskip-\ALG@thistlm}
\makeatother
\algdef{SE}[DOWHILE]{Do}{doWhile}{\algorithmicdo}[1]{\algorithmicwhile\ #1}

\graphicspath{ {figures/} }

\usepackage{arydshln}
\newcolumntype{x}[1]{>{\centering\arraybackslash\hspace{0pt}}p{#1}}
\newcolumntype{M}[1]{>{\centering\arraybackslash}m{#1}}
\newcolumntype{L}[1]{>{\raggedright\arraybackslash} m{#1} }

\usepackage{titlesec}

\begin{document}

\title{\LARGE \bf
{Wild-Places: A Large-Scale Dataset for Lidar Place Recognition \\ in Unstructured Natural Environments}
}

\author{Joshua Knights$^{*,1,2}$, Kavisha Vidanapathirana$^{*,1,2}$, Milad Ramezani$^1$, \\ Sridha Sridharan$^2$, Clinton Fookes$^2$, Peyman Moghadam$^{1,2}$ 
\thanks{
$^*$ Equal Contribution}

\thanks{
$^1$ Robotics and Autonomous Systems Group, DATA61, CSIRO, 
Australia. 
E-mails: {\tt\footnotesize \emph{
firstname.lastname
}@data61.csiro.au}}
\thanks{
$^{2}$ School of Electrical Engineering and Robotics, Queensland University of Technology (QUT), Brisbane, Australia.
}
}

\twocolumn[{%
\renewcommand\twocolumn[1][]{#1}%
\maketitle
\begin{center}
    \centering
    \captionsetup{type=figure}
    \includegraphics[width=\textwidth]{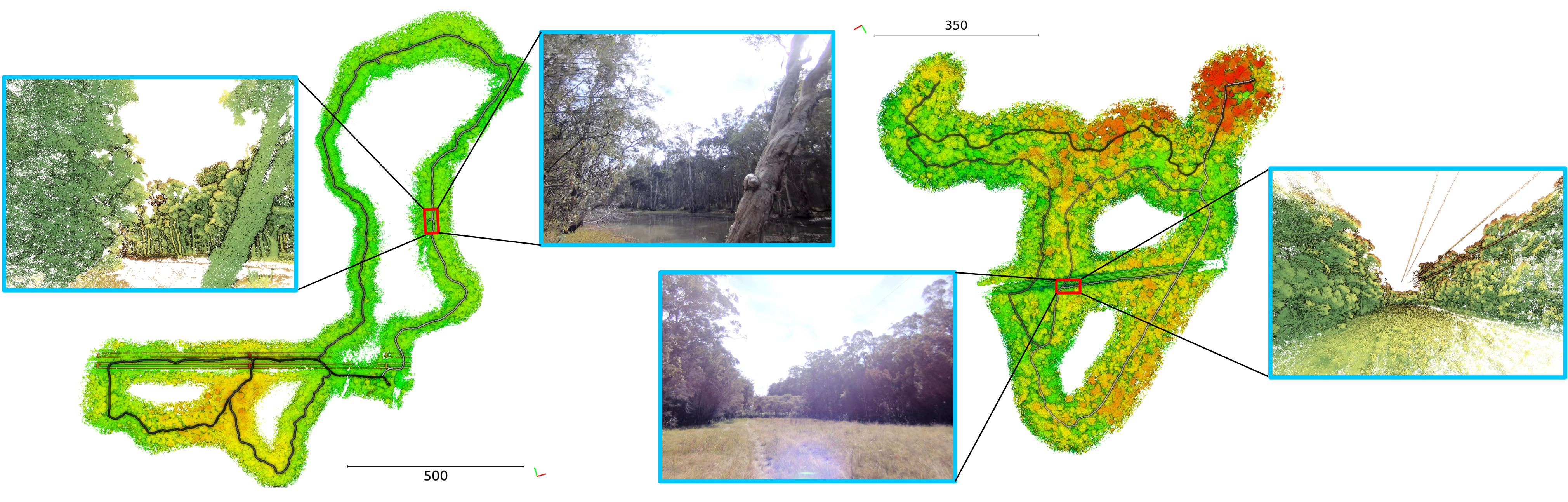}
    
    \captionof{figure}{Bird's eye view of the global map for two sequences from the Wild-Places dataset. The colour of the bold line in the figures represents the trajectory of the lidar sensor payload through the environment over time, while the colourisation of the global pointcloud is based off of the height of the individual points.  In the zoomed in regions, the boxes represent the camera image of the location and a corresponding point cloud for visualisation.} %
    \label{fig:hero}
    
\end{center}%
}]

\input{chapters/abstract}

\input{chapters/introduction-new}

\input{chapters/relatedworks}

\input{chapters/hardware.tex}

\input{chapters/dataset.tex}

\input{chapters/benchmarking.tex}

\input{chapters/experiments}

\input{chapters/conclusion.tex}

\section*{Acknowledgments}
The authors gratefully acknowledge funding of the project by the CSIRO's Machine Learning and Artificial Intelligence (MLAI) FSP. This work would not be possible without support from members of the CSIRO Robotics and Autonomous Systems including Brett Wood, Dennis Frousheger, Gavin Catt, Fred Pauling, Dave Haddon and Mark Cox.

\balance{}

{\small
        \bibliographystyle{IEEEtran}
        \bibliography{ref}
}

\end{document}

%% file: chapters/abstract.tex
\renewcommand{\thefootnote}{\fnsymbol{footnote}}

\footnotetext[0]{\vspace{0cm}$^*$ Equal contribution} 
\footnotetext[0]{\vspace{0cm}$^1$ Robotics and Autonomous Systems Group, DATA61, CSIRO, Australia.  E-mails: {\tt\footnotesize \emph{firstname.lastname}@data61.csiro.au}}
\footnotetext[0]{\vspace{0cm}$^2$ School of Electrical Engineering and Robotics, Queensland University of Technology (QUT), Australia. E-mails: {\tt\footnotesize \emph\{s.sridharan, c.fookes\}@qut.edu.au}} %

\begin{abstract}
Many existing datasets for lidar place recognition are solely representative of structured urban environments, and have recently been saturated in performance by deep learning based approaches. Natural and unstructured environments present many additional challenges for the tasks of long-term localisation but these environments are not represented in currently available datasets. 
To address this we introduce Wild-Places, a challenging large-scale dataset for lidar place recognition in unstructured, natural environments.  Wild-Places contains eight lidar sequences collected with a handheld sensor payload over the course of fourteen months, containing a total of 63K undistorted lidar submaps along with accurate 6DoF ground truth.  This dataset contains multiple revisits both within and between sequences, allowing for both intra-sequence (\ie~loop closure detection) and inter-sequence (\ie~re-localisation) tasks.  We also benchmark several state-of-the-art approaches to demonstrate the challenges that this dataset introduces, particularly the case of long-term place recognition due to natural environments changing over time.  Our dataset and code is available at \href{https://csiro-robotics.github.io/Wild-Places}{https://csiro-robotics.github.io/Wild-Places}

\end{abstract}

\renewcommand*{\thefootnote}{\arabic{footnote}}

%% file: chapters/introduction-new.tex
\section{Introduction}
\label{sec:intro}

\begin{table*}[t]
\begin{center}
\begin{tabular}{llcccll}
\hline
\multicolumn{1}{c}{\textbf{Dataset}} & \multicolumn{1}{c}{\textbf{Environment}} & \multicolumn{1}{c}{\begin{tabular}[c]{@{}c@{}}\textbf{Intra-sequence} \\ \textbf{revisits}\end{tabular}} & \multicolumn{1}{c}{\begin{tabular}[c]{@{}c@{}}\textbf{Inter-sequence} \\ \textbf{revisits}\end{tabular}} & \multicolumn{1}{c}{\begin{tabular}[c]{@{}c@{}}\textbf{Long term}  \\ \textbf{revisits}\end{tabular}} & \multicolumn{1}{c}{\begin{tabular}[c]{@{}c@{}}\textbf{Avg. points}\\ \textbf{per submap}\end{tabular}}  & \begin{tabular}[c]{@{}l@{}}\textbf{Distance} \\(km) \end{tabular} \\ \hline
KITTI~\cite{Geiger2013VisionMR} & Urban& \cmark& - & - & 120K & 44 \\
KITTI360~\cite{kitti360}& Urban& \cmark& \cmark & -  & 120K  & 67 \\
NCLT~\cite{carlevaris2016nclt} & Urban & \cmark & \cmark & \cmark & 50k  & 150  \\
Oxford~\cite{RobotCarDatasetIJRR} & Urban  & \cmark & \cmark & \cmark & 4K$^*$  & 1000 \\
Oxford-Radar~\cite{barnes2020oxford} & Urban  & \cmark & \cmark & -  & 21K  & 280 \\
Complex Urban~\cite{jeong2019complexurban} & Urban  & \cmark & - & - & 20K  & 190  \\
MulRan~\cite{Kim2020MulRanMR} & Urban  & \cmark & \cmark & -  & 64K  & 123  \\
Apollo-Southbay~\cite{lu2019l3} & Urban  & \cmark  & \cmark  & - & 120K  & 398  \\
\hline
Rellis 3D~\cite{jiang2021rellis3d} & Natural  & -  & -  & -  & 13K  & 1.5  \\
ORFD~\cite{min2022orfd} & Natural  & -  & -  & -  & 72K  & 3  \\
\textbf{Wild-Places (Ours)} & Natural  & \cmark & \cmark & \cmark & 300K  & 33  \\ 
\hline
\end{tabular}
\caption{Comparison of public lidar datasets. The top half of the table shows the most popular lidar datasets used for large-scale localisation evaluation. The bottom half shows public lidar datasets which contain only natural and unstructured environments. Wild-Places is the only dataset that satisfies both of these criteria.  We define long-term revisits here as a time gap greater than 1 year. $^*$ Post-processed variation introduced in \cite{Uy2018PointNetVLADDP}. }
\label{tab:comparison_table}
\end{center}
\vspace{-0.75cm}
\end{table*}

Lidar Place Recognition is an essential element for safe and reliable deployment of robots for long-term simultaneous localisation and mapping (SLAM) in challenging environments \cite{zhang2022hilti, park2021elasticity}. Advances in this field have been accelerated by large-scale lidar datasets and benchmarks such as KITTI~\cite{Geiger2013VisionMR}, MulRan~\cite{Kim2020MulRanMR} and Oxford RobotCar~\cite{RobotCarDatasetIJRR}, which are exclusively representative of urban, on-road environments. 
There are increasing demands for development of robotic solutions for unstructured natural environments for a range of applications including agriculture, environmental monitoring, conservation and search and rescue \cite{tartandrive, hudson2022heterogeneous}. 
Natural environments are often characterised by highly irregular and unstructured terrain, narrow trails surrounded by dense vegetation with overhanging branches, which pose significant challenges to existing state-of-the-art place recognition methods, particularly in the case of long-term place recognition due to dynamic shifts in the environment over time.  

Consequently, there is a great need for large-scale lidar datasets for long-term place recognition and localisation tasks in unstructured, natural environments. Recently several off-road datasets have been introduced~\cite{jiang2021rellis3d, wigness2019rugd, leyva2019tb, valada2016deepmultispectraldataset, tartandrive} to address this demand; however, while these datasets provide rich training data for semantic segmentation~\cite{jiang2021rellis3d, wigness2019rugd, min2022orfd} or vehicle dynamics  prediction~\cite{tartandrive} tasks, they are not suitable for training and evaluation of long-term place recognition methods. For the lidar place recognition task, a dataset should ideally contain multiple large and continuous traverses of many kilometers in natural environments that include loops and revisits both within and between sequences to allow evaluation of both intra-sequence place recognition (similar to the loop closure detection problem) and inter-sequence place recognition (similar to the re-localisation problem). To this end we created Wild-Places, a large-scale dataset for long-term lidar place recognition in unstructured, natural environments.  

The dataset is collected in national parks in Brisbane Australia, by a human operator carrying a handheld sensor payload allowing for recording of dense forest environments that are inaccessible to vehicular or robotic platforms. The data collection campaign was conducted in 2 forest trails and consists of more than $33$ kilometers of traversal over the course of $14$ months, generating roughly $63$K lidar submaps.

The contributions of this dataset are as follows: 
 \begin{itemize}
     \item We introduce the first large-scale lidar dataset, collected with a handheld sensor payload, for long-term place recognition in unstructured, natural environments.  %
     \item We show that our dataset is suitable for intra-sequence and inter-sequence place recognition tasks, and establish training and testing splits for benchmarking.  %
     \item We benchmark several state-of-the-art place recognition methods to demonstrate the challenging scenarios presented by Wild-Places dataset. These results demonstrate the challenges posed by unstructured natural environments for these tasks, and help identify areas of research which require additional attention.
     
 \end{itemize}

%% file: chapters/relatedworks.tex
\section{Related Work}
\label{sec:relwork}

Progress in the field of lidar place recognition has been driven by the availability of large-scale datasets for benchmarking and evaluation, of which we provide an overview in~\tabref{tab:comparison_table}.  The top half of the table shows the most popular lidar datasets used for large-scale localisation evaluation.  Datasets such as KITTI \cite{Geiger2013VisionMR}, Oxford\cite{RobotCarDatasetIJRR}, NCLT\cite{carlevaris2016nclt} and Complex Urban\cite{jeong2019complexurban}  have been historically popular benchmarks, and recently several new datasets such as KITTI-360 \cite{kitti360}, Oxford Radar RobotCar \cite{barnes2020oxford} and MulRan \cite{Kim2020MulRanMR} have either extended on existing datasets with new traversals and modalities or introduced new place recognition scenarios in multiple urban environments.  Several other lidar datasets \cite{caesar2020nuscenes,chang2019argoverse,wilson2021argoverse,steder2010frieburgdataset,sun2020scalabilitywaymo,pandey2011ford,peynot2010marulan,ramezani2020newer} 
are not included in the table as they are currently not used for evaluation in recent place recognition methods due to a combination of the following reasons: small spatial scale, low amount of loops and/or the lack of reverse revisits.

However even with these recent additions and extensions, recent state-of-the-art place recognition methods are beginning to saturate benchmarks on these urban datasets \cite{vidanapathirana2021locus, komorowski2021egonn, vidanapathirana2021logg3d, cattaneo2022lcdnet, komorowski2022improving, Vidanapathirana2023SpectralGV}. In the recently held `General Place Recognition: City-scale UGV Localization' challenge at ICRA 2022, top-performing methods reached a top-1 Recall of above $99$\% during the first round of the competition.  In addition, we observe that recent learning-based methods trained on urban datasets transfer extremely well to other unseen urban datasets~\cite{komorowski2021egonn, cattaneo2022lcdnet}, indicating the lack of a significant domain shift between urban, on-road environments.  We hypothesise that the distinct structural elements present in urban environments are now fairly trivial for large deep neural networks to represent and discriminate. 
Consequently, in order to promote further progress in the field, datasets that cover larger and more diverse environments are needed, such as those presented in natural environments.

Beyond the scope of lidar place recognition, recently there has been an increase in focus for scene understanding tasks in natural and unstructured environments~\cite{tartandrive, shaban2022semantic, baril2021kilometer}.  Natural environments do not have many of the structural cues present in on-road, urban environments (\eg~flat ground, rigid scene elements), increasing the challenge of 3D scene understanding tasks in these settings.  RUGD~\cite{wigness2019rugd} , Freiburg Forest~\cite{valada2016deepmultispectraldataset}, TB-Places~\cite{leyva2019tb} and TartanDrive~\cite{tartandrive} are all large datasets which provide data for different modalities and tasks in natural environments, but do not contain any lidar data.  %

The bottom half of Table \ref{tab:comparison_table} summarises lidar datasets collected in natural environments. ORFD\cite{min2022orfd} introduces a  dataset for off-road free-space detection, which covers different off-road scenes (woodland, farmland, grassland and countryside) in different weather and lighting conditions. 
 ORDF consists of 30 sequences where each sequence only contains a traversal of roughly 100$m$, which is too small for the evaluation of large-scale localisation. The RELLIS-3D dataset \cite{jiang2021rellis3d} provides RGB and lidar annotations for the task of multi-modal semantic segmentation in natural environments. 
 RELLIS-3D consists of 5 sequences each without loops where each sequence is only a couple of hundred meters (total of 1500 meters across all sequences), which is not sufficient for evaluation of place-recognition.
 In order to address these gaps our dataset contains 33 kilometres of traversal over 14 months in two different environments, and supports training and evaluation for intra-sequence revisits and both long and short-term inter-sequence revisits.

\begin{table}[b!]
\caption{Sensors specification.}
\label{tab:sensors}
\centering
\begin{tabular}{l l l l} 
\hline
 \textbf{Sensor} & \textbf{Model} & \textbf{Rate} (Hz) & \textbf{Specifications}\\
 \hline
  Lidar & VLP-16 & 20 &  16 Channels\\
  &  &  &  120 m Range\\
  Camera x4 & e-CAM130A CUXVR & 15 &  $94.9^{\circ}$H  FOV \\
  &  &  &  $71.2^{\circ}$V FOV \\
  IMU & 3DM-CV5-25 & 100 &  9 DoF \\
  DC Motor &  & 0.5 & Brushless  \\
  Encoder &  & 100 & PPS Synched  \\
 \hline
\end{tabular}
\vspace{-2mm}
\end{table}

%% file: chapters/hardware.tex
\section{Hardware Setup}
\label{sec:hardward}

\begin{figure}
    \centering
    \includegraphics[width=1.0\columnwidth]{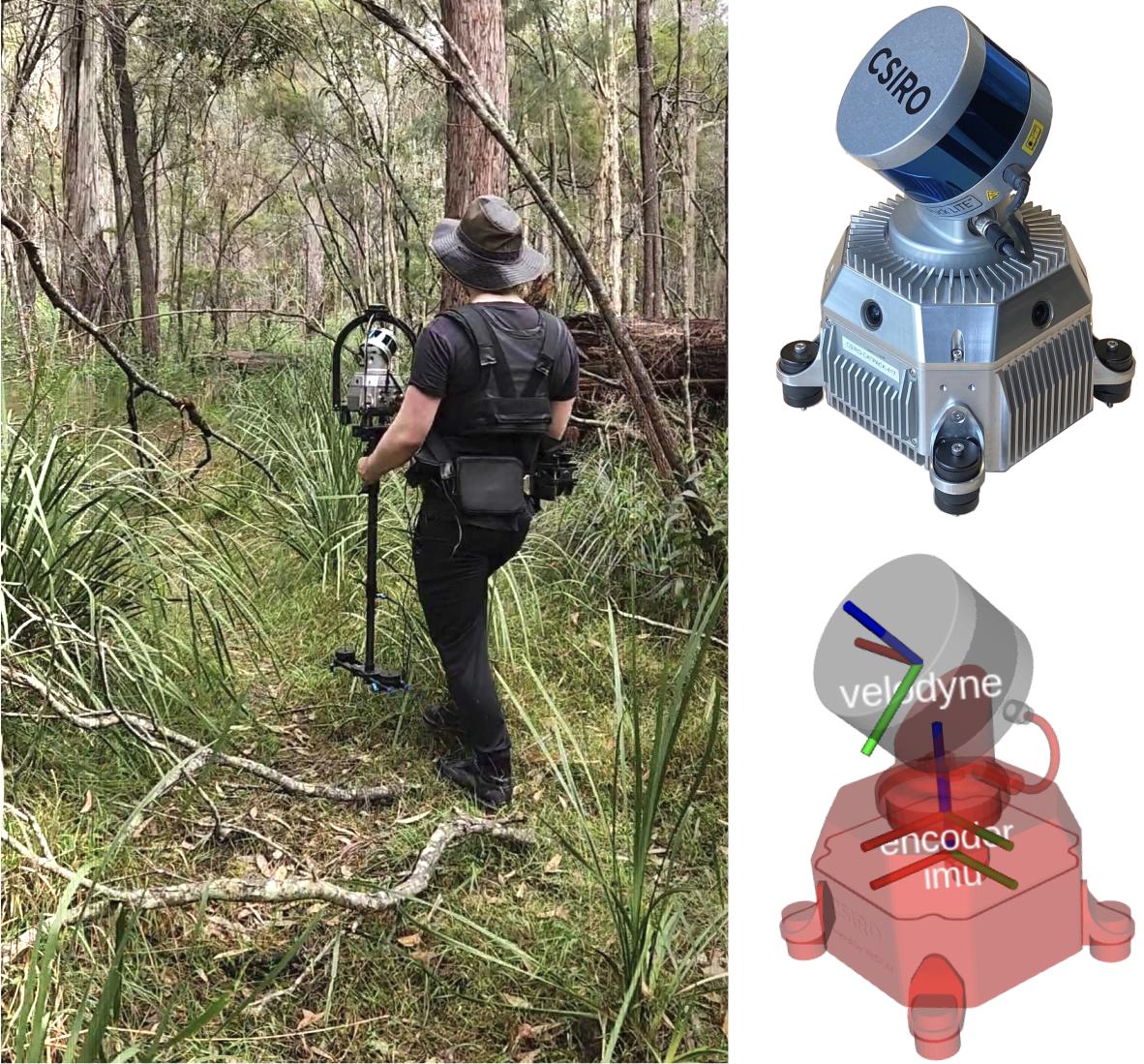}
    \caption{\textbf{Left}: Image of sequence V-04 data collection campaign depicting the dense forest trails which are not accessible using vehicle or robotic platforms. \textbf{Right}: The sensor payload consists of a spinning lidar sensor mounted at an angle of $45^{\degree}$ to maximise field of view, a motor, encoder, an IMU, and four cameras. %
    }
    \label{fig:catpack}
\end{figure}

Our handheld sensor payload
consists of a Velodyne VLP-16 lidar scanner which is mounted on a brushless DC motor, spinning around the $z$ axis at 0.5 Hz, a Microstrain 3DM-CV5-25 9-DoF IMU and a Nvidia Jetson AGX Xavier, as shown in~\figref{fig:catpack}. 
A timing and control board interfaces to the encoder and controls the motor and interfaces with the Xavier via ROS. 
Pulse Per Second (PPS) is used on the device for time synchronisation between the sensors. 
The sensor payload also contains four cameras for visual perception.
Since the VLP-16 has a limited vertical field of view (30\degree), it is attached to the DC motor with a 45\degree~inclination and rotated about an external axis. This design allows lidar scans of 120\degree~vertical FoV suitable to map features such as trees from top to bottom.  
~\figref{fig:catpack} shows how the sensor payload was carried using a harness. \tabref{tab:sensors} gives an overview of the specifications of the sensors.

\vspace{-2mm}

%% file: chapters/dataset.tex
\section{Data Colllection}
\label{sec:dataset}

The Wild-Places dataset consists of undistorted lidar sequences along with accurate 6DoF ground truth collected in unstructured, natural environments
by a human operator using a handheld sensor payload depicted in~\figref{fig:catpack}.  We first describe the properties of the lidar sequences and visited environments in our dataset before detailing the process of generating point cloud submaps for training and evaluation. 

\subsection{Environment and Sequence Details}
\begin{figure*}
    \centering
    \captionsetup{type=figure}
    \includegraphics[width=\textwidth]{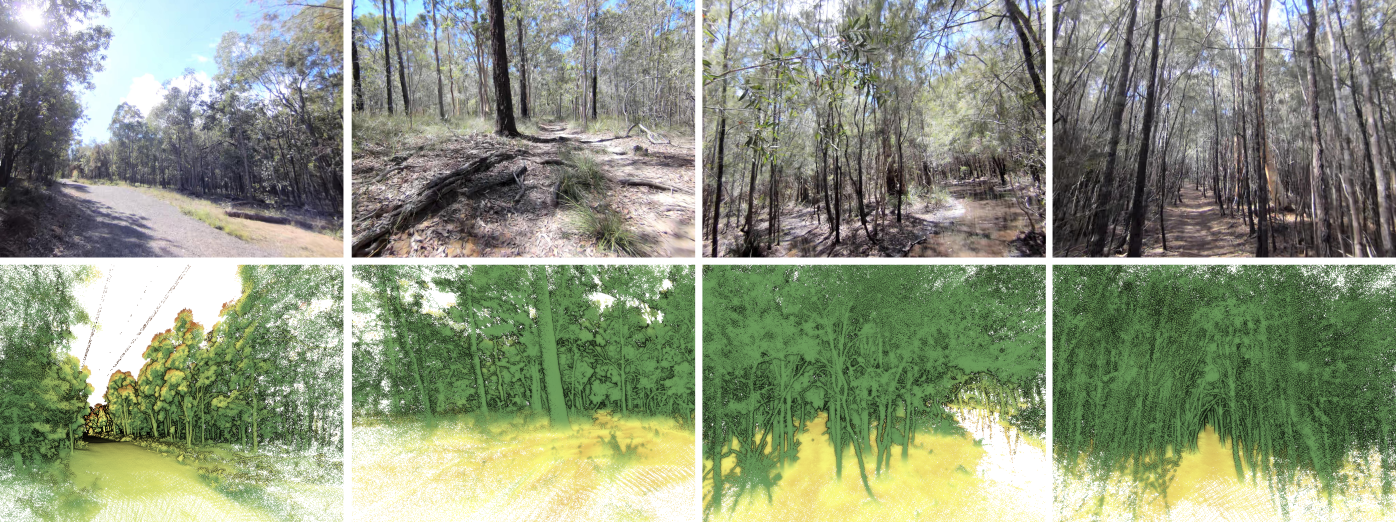}
    
    \captionof{figure}{Examples of structural diversity within the Wild-Places dataset. The top row shows the RGB images and the bottom shows the corresponding point clouds visualised from the camera view point. From left to right, we see wide trail surrounded by tall trees, sparse forest with roots and logs, small bodies of water/mud that show no lidar returns and dense vegetation with repetitive tree structures.} 
    \label{fig:diversity}
    
\vspace{-0.5cm}
\end{figure*}

Our dataset consists of eight lidar sequences collected in two environments, which we refer to as Venman and Karawatha.  \figref{fig:diversity} demonstrates the in-sequence diversity of these environments, containing both sparsely and densely vegetated areas with a variety of terrain characteristics. \tabref{tab:seq_details}  details the properties of each sequence in the dataset.  We use V-XX and K-XX as shorthand to refer to Sequence XX on the Venman and Karawatha environments, respectively.  Sequence $01$ and $02$ for both environments were collected on the same day, with Sequence $02$ following the reverse route of Sequence $01$.  Sequence $03$ for both environments were collected six months later and follow extended alternative routes, while Sequences $04$ were collected 14 months after Sequence $01$ and follow the same routes as Sequences $01$.  
Each sequence contains multiple in-sequence revisits in both forward and reverse directions.  These properties allow for our lidar sequences to be used in evaluation involving reverse revisits, intra-run loop closure detection, and inter-run re-localisation over both short and long term revisits between sequences.

\begin{table}[b]
    \centering
    \caption{Wild-Places dataset.}
    \begin{NiceTabular}{cccccc}
        \hline 
        \Block{1-2}{\textbf{Sequence}} & & \textbf{Date} & \textbf{Length} & \textbf{Duration} & \textbf{Submaps}\\
        \hline 
        \Block{4-1}{Venman} 
        & 01 & June 2021  & 2.64 km & 39m & 4706 \\
        & 02 & June 2021 & 2.64 km & 38m & 4557 \\
        & 03 & Dec 2021 & 4.59 km & 1h 11m & 8470\\
        & 04 & Aug 2022 & 2.81 km & 48m & 5739 \\
        \hline 
        \Block{4-1}{Karawatha} 
        & 01 & June 2021 & 5.14 km & 1h 14m & 8817 \\
        & 02 & June 2021 & 5.66 km & 1h 24m & 10075 \\
        & 03 & Dec 2021 & 6.27 km & 2h 7m & 15150 \\
        & 04 & Aug 2022 & 3.17 km & 48m &  5805 \\
        \hline 
        Total & 8 & 14 months & 32.87 km & 8h 51m & 63319 \\
        \hline 
    \end{NiceTabular}
    
    \label{tab:seq_details}
    \vspace{-0.5cm}
\end{table}

\subsection{Ground Truth}
\label{sec:GT}
To accurately generate the map of each sequence, we use the Wildcat SLAM~\cite{ramezani2022wildcat} system. Wildcat SLAM employs a continuous-time trajectory representation to integrate asynchronous IMU and lidar measurements in a sliding window to deal with lidar distortion caused by motion. 
An odometry module estimates a robust pose by intergrating constraints on IMU measurements along with correspondences between surfels generated by segmenting the lidar points based on their location and time of capture. 
To mitigate drift over time, pose graph optimisation is used in which submaps spawned from odometry are adjusted relative to one another along with loop closure constraints, resulting in a globally consistent map. 
Wildcat has demonstrated robust mapping of large outdoor environments reporting mean errors of ~0.3 meters when compared with ground truth obtained by accurate surveying\cite{ramezani2022wildcat}.
In addition, \cite{zhang2022hilti} recently reported the SLAM results from the \textit{Hilti SLAM Challenge 2022} in which Wildcat SLAM achieved the highest scoring solution compared to the other lidar-inertial SLAM techniques. Hence we generated accurate intra-sequence ground truth for this dataset.   

\begin{figure}[t]
    \centering
    \begin{subfigure}[t]{0.49\columnwidth}
        \frame{\includegraphics[width=1\textwidth]{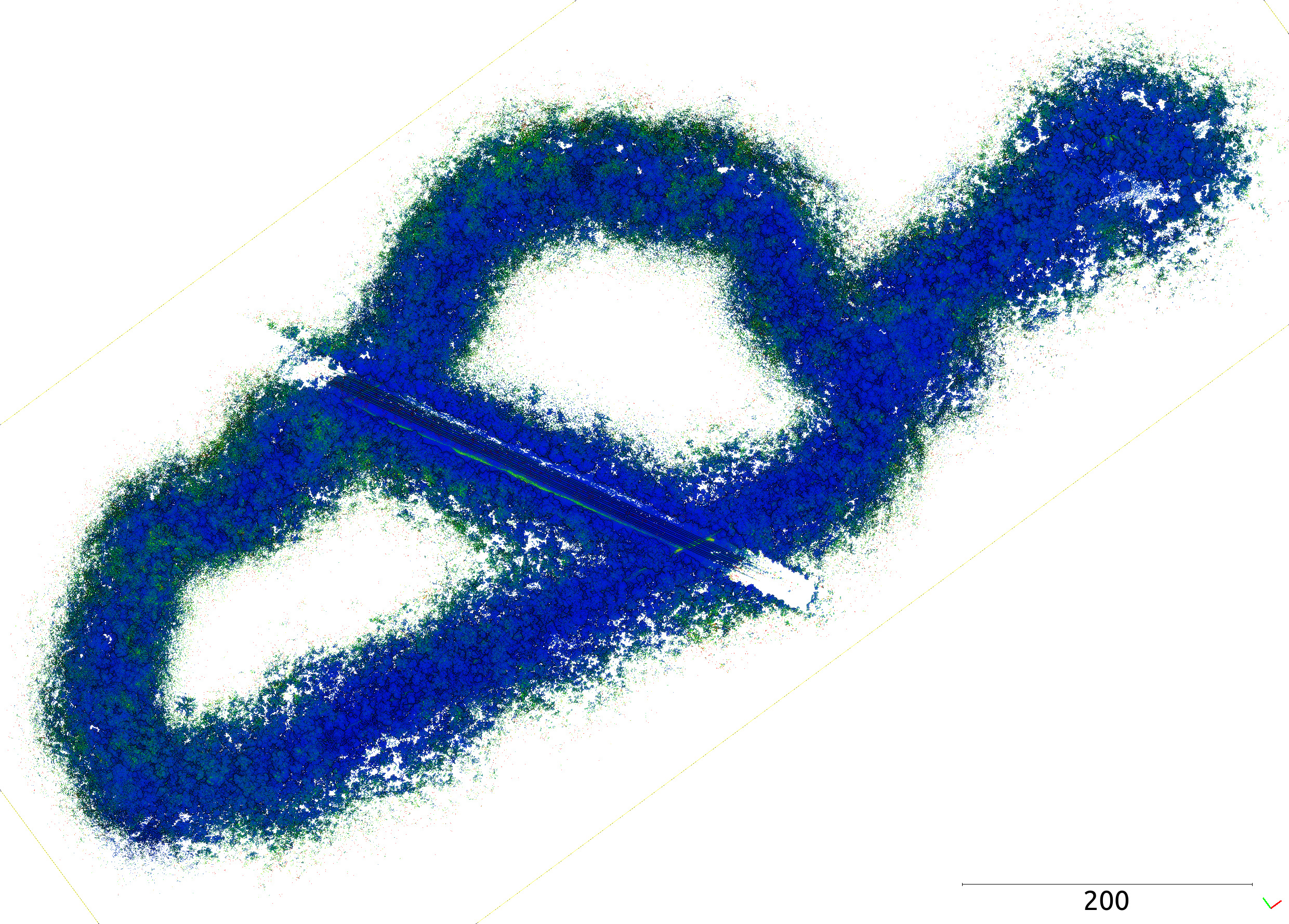}}
    \end{subfigure}%
    \hfill 
    \begin{subfigure}[t]{0.49\columnwidth}
        \frame{\includegraphics[width=1\textwidth]{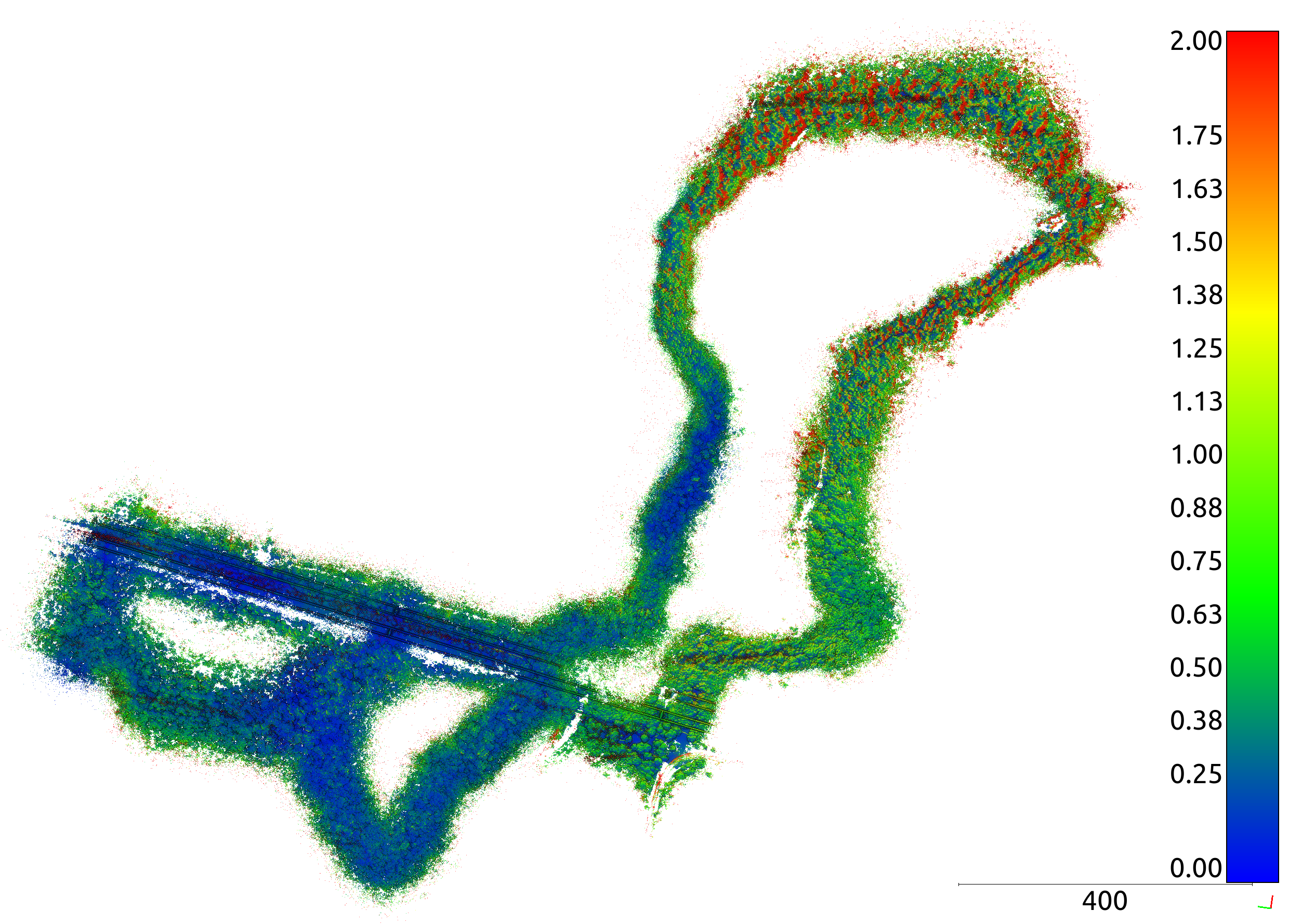}}
    \end{subfigure}
    \caption{Visualisation of point-to-point distances after registration between the global maps of two sequences of Venman (left) and Karawatha (right).  Colour bar shows the distance between correspondences.} 
    \label{fig:gt}
    \vspace{-0.5cm}
\end{figure}

For inter-sequence registration, we align the global maps of sequences relative to one another utilising Iterative Closest Point (ICP) and remove outliers using the M-estimator Sample Consensus (MSAC) method~\cite{torr2000mlesac}.~\figref{fig:gt} demonstrates the point-to-point comparison between two sequences of Venman (left) and Karawatha (right) to visualise the quality of the registration between the sequences. In both examples above $95\%$ of correspondences have error less than 0.95m, which we believe to be sufficiently accurate for training and evaluating inter-sequence place recognition tasks. %

\subsection{Submap Generation}
To generate our point cloud submaps for training and evaluation, we first generate the global map and trajectory for each sequence as outlined in \secref{sec:GT}. Every 0.5 seconds through the trajectory we sample a point cloud submap with a diameter of 60 metres centered on the current position of the sensor in the global map, which we record alongside the 6 Degree-of-Freedom (DoF) pose of the sensor.  To allow our dataset to be used for investigating intra- and inter-sequence place recognition, we also only sample points within a one second window of the corresponding timestamp for the submap, ensuring there are no shared points between submaps for in-sequence revisits.  Across all eight sequences we generate a total of $63,319$ point cloud submaps with over 300K points per submap on average. To the best of our knowledge, this is the largest and highest density lidar dataset in natural environments.

\begin{figure}[ht]
    \centering
    \begin{subfigure}[t]{0.5\columnwidth}
         \centering
         \includegraphics[width=1\textwidth]{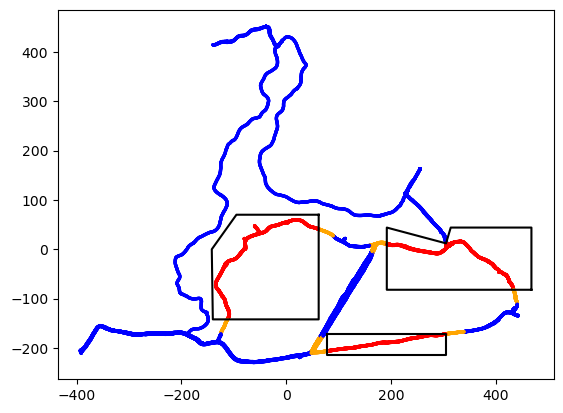}
         \label{fig:Venman_GT}
     \end{subfigure}%
     \hfill
     \begin{subfigure}[t]{0.5\columnwidth}
         \centering
         \includegraphics[width=1\textwidth]{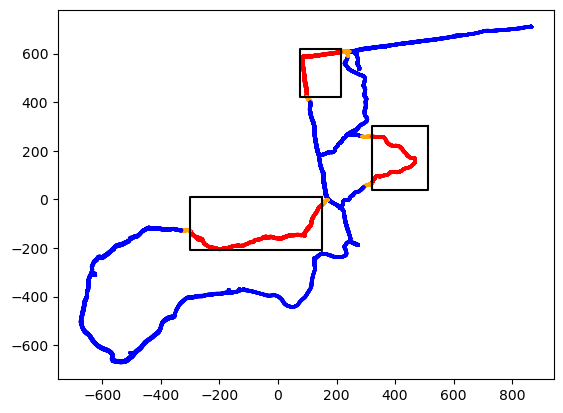}
         \label{fig:Karawatha_GT}
     \end{subfigure}%
     \vspace{-0.5cm}
    \caption{Training and testing splits visualised on the trajectories of all sequences on Venman (left) and Karawatha (right).  Regions highlighted in red are used as queries during test time, regions highlighted in blue are used during training, and regions highlighted in orange are withheld to prevent information leaking between the queries and the training set.}
    \label{fig:splits}
    \vspace{-0.5cm}
    
\end{figure}

\subsection{Data Format}

The point clouds in each sequence are stored in the format $ <timestamp.pcd> $, where each .pcd file corresponds to a lidar submap containing the $x, y, z$ values of its points. The trajectory of each sequence is stored in a $ <poses.csv> $ file which contains the timestamp and the 6DoF global pose of each submap, where the pose information is stored in $x,y,z$ for position and $qx,qy,qz,qw$ quaternion for orientation.

%% file: chapters/benchmarking.tex
\section{Benchmarking}
\label{sec:benchmarking}

In this section we establish the training and testing splits used for our benchmarking of the Wild-Places dataset, as well as a summary of the tasks and metrics used for evaluation.

\subsection{Training \& Evaluation Splits}

To form our training and evaluation splits, we first align the submap poses for each sequence using the transforms between the global maps derived in \secref{sec:GT}.  We then split the submaps generated for each sequence into two disjoint sets as shown in Figure \ref{fig:splits}.  The regions highlighted in red are withheld during training and used as queries for evaluating performance at test time for inter-sequence evaluation.  The regions highlighted in orange are not used as queries, but are also withheld during training in order to prevent leakage between the training and testing sets. Our splits are selected to cover a diverse set of scenes within each environment while containing as few human-made structures as possible in order for our evaluation to focus on the challenges presented by the unstructured, natural environments.  In addition, we only use the sequences $01$ and $02$ for training for two reasons: (1) we reserve sequences $03$ and $04$ for intra-sequence loop closure detection evaluation, and (2) since sequences $03$ and $04$ are collected 6 months and 14 months, respectively, after sequences $01$ and $02$, they are used to test the challenges posed by long-term changes which are abundant in natural environments.  This configuration gives us a total of $19,096$ training submaps and $16,037$ query submaps across all eight sequences.

\subsection{Inter-Sequence Evaluation}
We first conduct inter-sequence evaluation which is similar to the re-localisation problem. At test time, we use the previously discussed withheld regions as queries and every submap from other sequences in the same environment as a database.  
We consider a submap successfully localised if the retrieved submap from the database is within three metres of the query. We evaluate performance using the metrics of top-1 Recall (R1) and mean reciprocal rank (MRR) on each environment, with the latter being defined as  $MRR=\frac{1}{N}\sum_{i=1}^{N} \frac{1}{rank_i}$ where $rank_i$ is the rank of the first retrieved positive match to query submap $i$. For each sequence, we evaluate against the 3 other sequences in the same environment, giving a total on 12 evaluations per environment. The final R1 and MRR values used to compare different methods are the means of the respective R1 and MRR values over these 12 evaluations. 

\subsection{Intra-Sequence Evaluation}
We also conduct intra-sequence evaluation on sequences V-$03$, V-$04$, K-$03$ and K-$04$. This is akin to the loop-closure detection problem where each submap in a sequence acts as a query to match against  previously seen point clouds in the same sequence. Previous entries adjacent to the query by less than $ 600 s $ time difference are excluded from the search to avoid matching to the same instance. We consider a submap successfully localised if the retrieved submap is within three metres. 
Compared to the inter-sequence re-localisation evaluation, queries in intra-sequence loop-closure evaluation are not guaranteed to be revisits and can be true-negatives/false-positives. To account for this we opt to report the maximum F1 score (F1) along with the top-1 Recall (R1) for the evaluation of different methods.

%% file: chapters/experiments.tex
\section{Experimental Evaluation}
\label{examples}
In this section, we briefly cover the state-of-the-art methods used in our benchmarking before presenting and discussing the results of the experimental setup described in~\ref{sec:benchmarking}. 
\begin{table*}[t!]
    \centering
    \caption{Benchmarking results for intra and inter-sequence place recognition performance on the Wild-Places dataset.}
    \begin{NiceTabular}{cw{c}{0.45cm}w{c}{0.45cm}w{c}{0.45cm}w{c}{0.45cm}w{c}{0.45cm}w{c}{0.45cm}w{c}{0.45cm}w{c}{0.45cm}w{c}{0.45cm}w{c}{0.45cm}|w{c}{0.45cm}w{c}{0.45cm}w{c}{0.45cm}w{c}{0.45cm}w{c}{0.45cm}w{c}{0.45cm}}
        \hline
        \Block{3-1}{\textbf{Method}} & \Block{1-10}{\textbf{Intra-Sequence}} & & & & & & & & & & \Block{1-6}{\textbf{Inter-Sequence}} & & & & & \\ 
        \cmidrule(lr){2-11} \cmidrule(lr){12-17}
        &\Block{1-2}{\textbf{V-03}} &  & \Block{1-2}{\textbf{V-04}} & & \Block{1-2}{\textbf{K-03}} & & \Block{1-2}{\textbf{K-04}} & & \Block{1-2}{\textbf{Average}}
        &&\Block{1-2}{\textbf{Venman}} & & \Block{1-2}{\textbf{Karawatha}} & &   \Block{1-2}{\textbf{Average}}& \\
        \cmidrule(lr){2-3} \cmidrule(lr){4-5} \cmidrule(lr){6-7} \cmidrule(lr){8-9} \cmidrule(lr){10-11} \cmidrule(lr){12-13} \cmidrule(lr){14-15} \cmidrule(lr){16-17}
        &\textbf{F1} & \textbf{R1} & \textbf{F1} & \textbf{R1} & \textbf{F1} & \textbf{R1} & \textbf{F1} & \textbf{R1} & \textbf{F1} & \textbf{R1}
        &\textbf{R1} &  \textbf{MRR} & \textbf{R1} & \textbf{MRR} &\textbf{R1} &  \textbf{MRR} \\
        
        \hline
        ScanContext \cite{kim2018scan} &
        \red{01.01} & \red{03.31} & \red{13.00} & \red{28.92} & \red{13.22} & \red{31.42} & \red{32.26} & \red{70.99} & \red{14.87} & \red{33.66} &
        \red{33.98} & \red{64.67}  &\red{38.44}  & \red{67.90}  & \red{36.21} & \red{66.29}  \\
        \hline
        TransLoc3D \cite{Xu2021TransLoc3DP} & 
        \red{17.09} & \red{34.51} & \red{60.83} & \red{53.04} & \red{47.22} & \red{43.20} & \red{66.99} & \red{58.48} & \red{48.03} & \red{47.31} &
        \red{50.24} & \red{66.16}  &\red{46.08}  & \red{50.24}  & \red{48.16} & \red{58.20}  \\
        \hline
        MinkLoc3Dv2 \cite{komorowski2022improving} & 
        \red{49.78} & \red{49.94} & \red{\textbf{82.19}} & \red{71.61} & \red{51.33} & \red{50.97} & \red{80.00} & \red{71.18} & \red{65.83} & \red{60.93} &
        \red{75.77} & \red{84.87}  &\red{67.82}  & \red{79.21}  & \red{71.80} & \red{82.04}  \\
        \hline 
        LoGG3D-Net \cite{vidanapathirana2021logg3d} &
        \red{\textbf{53.94}} & \red{\textbf{62.40}} & \red{80.42} & \red{\textbf{72.47}} & \red{\textbf{64.30}} & \red{\textbf{64.05}} & \red{\textbf{84.54}} & \red{\textbf{80.26}} & \red{\textbf{70.80}} & \red{\textbf{69.80}} &
        \red{\textbf{79.84}} & \red{\textbf{87.33}}  &\red{\textbf{74.67}}  & \red{\textbf{83.68}}  & \red{\textbf{77.26}} & \red{\textbf{85.51}}  \\
        \hline
    \end{NiceTabular}
    
    \label{tab:benchmarking}
\end{table*}

\subsection{State-of-the-art Methods}
We benchmark four different approaches for lidar place recognition.  ScanContext \cite{kim2018scan} is a handcrafted approach which encodes the highest z-value of a point cloud split into bins from a bird's eye view, and is a widely used baseline for lidar place recognition in urban environments.  TransLoc3D \cite{Xu2021TransLoc3DP}, MinkLoc3Dv2 \cite{komorowski2022improving} and LoGG3D-Net \cite{vidanapathirana2021logg3d} are all state-of-the-art learning-based methods and have saturated performance on the popular urban lidar datasets.
TransLoc3D and MinkLoc3Dv2 achieve a top-1 Recall of $95.0$\% and $96.9$\% respectively on the commonly used Oxford RobotCar \cite{RobotCarDatasetIJRR} benchmark, and LoGG3D-Net achieves a mean maximum F1 score of $93.9$\% and $96.8$\% on the KITTI \cite{Geiger2013VisionMR} and MulRan \cite{Kim2020MulRanMR} benchmarks, respectively.   

\subsection{Results}
\begin{figure}[t]
    \centering
    \begin{subfigure}[t]{0.5\columnwidth}
         \centering
         \includegraphics[width=1\textwidth]{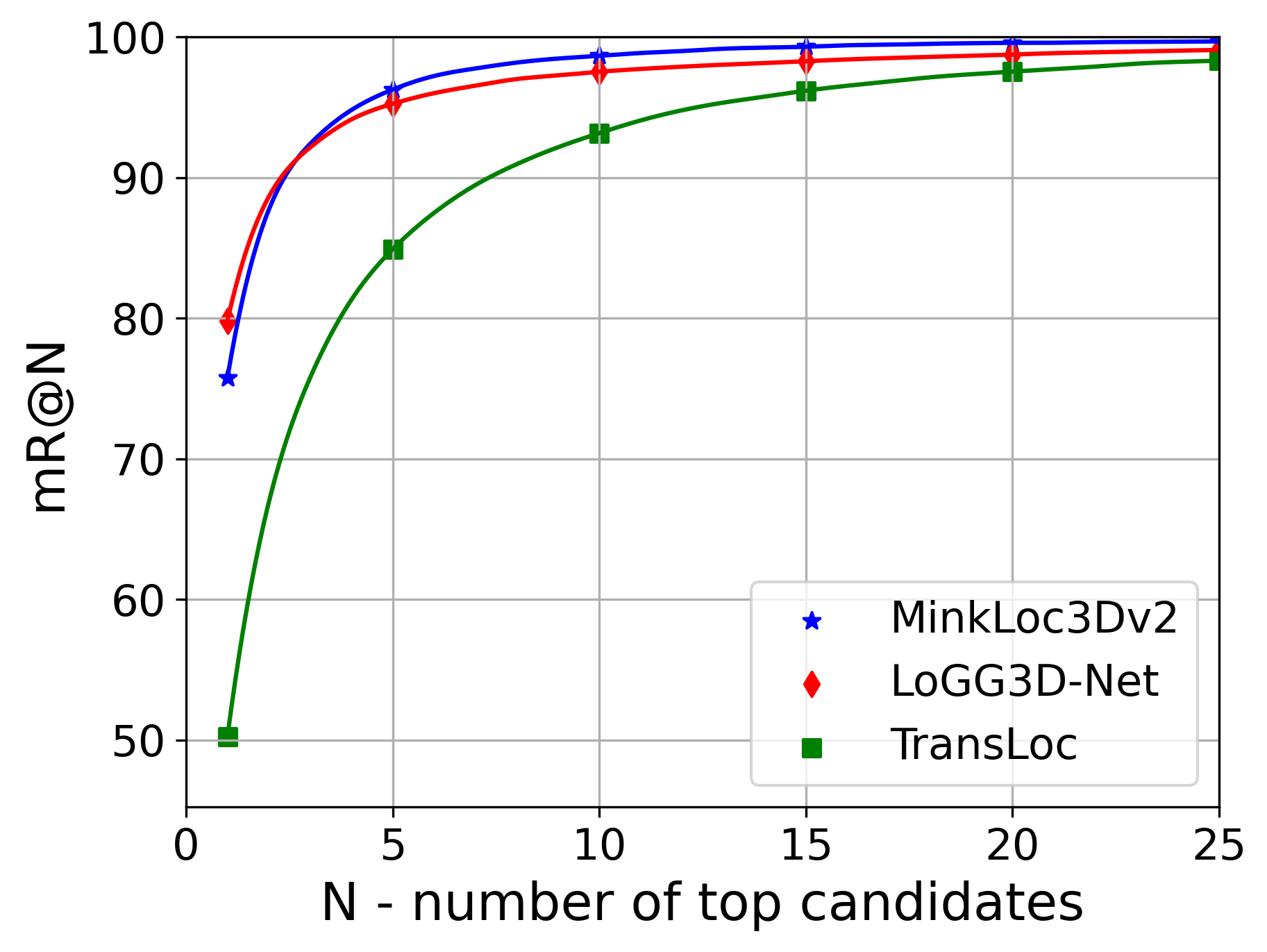}
         \caption{Venman}
         \label{fig:R@N_Venman}
     \end{subfigure}%
     \hfill
     \begin{subfigure}[t]{0.5\columnwidth}
         \centering
         \includegraphics[width=1\textwidth]{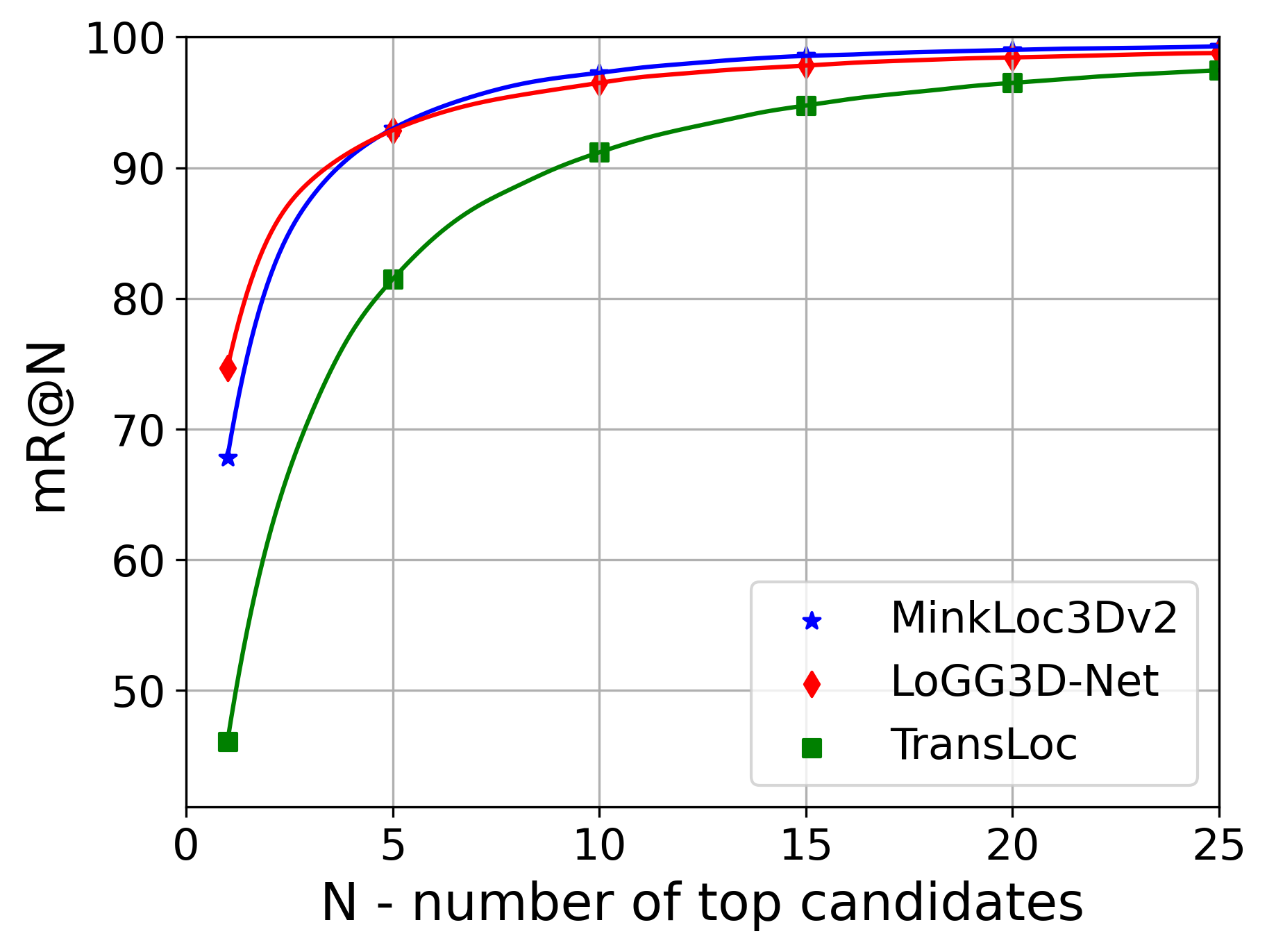}
         \caption{Karawatha}
         \label{fig:N_Karawatha}
     \end{subfigure}%
     \caption{Recall@N performance of learning-based approaches on the Venman and Karawatha environments.}
     \label{fig:r@N}
\end{figure}

Table \ref{tab:benchmarking} presents the results for both intra-sequence and inter-sequence evaluations.  We report metrics as outlined in Section \ref{sec:benchmarking} and visualise the top-N Recall curves for inter-sequence evaluation for learning-based methods in Figure \ref{fig:r@N}.  

\textbf{Intra-Sequence Evaluation: } For intra-sequence evaluation we observe that approaches which have been shown to saturate benchmark datasets in urban environments have significant room for improvement on our dataset, with a highest reported average max-F1 and top-1 Recall of only $70.8$\% and $69.8$\% compared to the mid-to-high $90$\% seen in urban environments. 
We observe that the sequence V-03 which has the highest amount of in-sequence reverse-revisits of all 8 sequences proves the most challenging for all methods. Conversely, K-04 which has barely any reverse revisits gives the highest performance for all methods. 
This drop in performance is particularly prominent in ScanContext \cite{kim2018scan} which we attribute to the strategy of encoding highest z-values being unsuited for environments dominated by relatively uninformative dense overhead canopy.

\begin{figure}[t]
    \centering
    \includegraphics[width=\columnwidth]{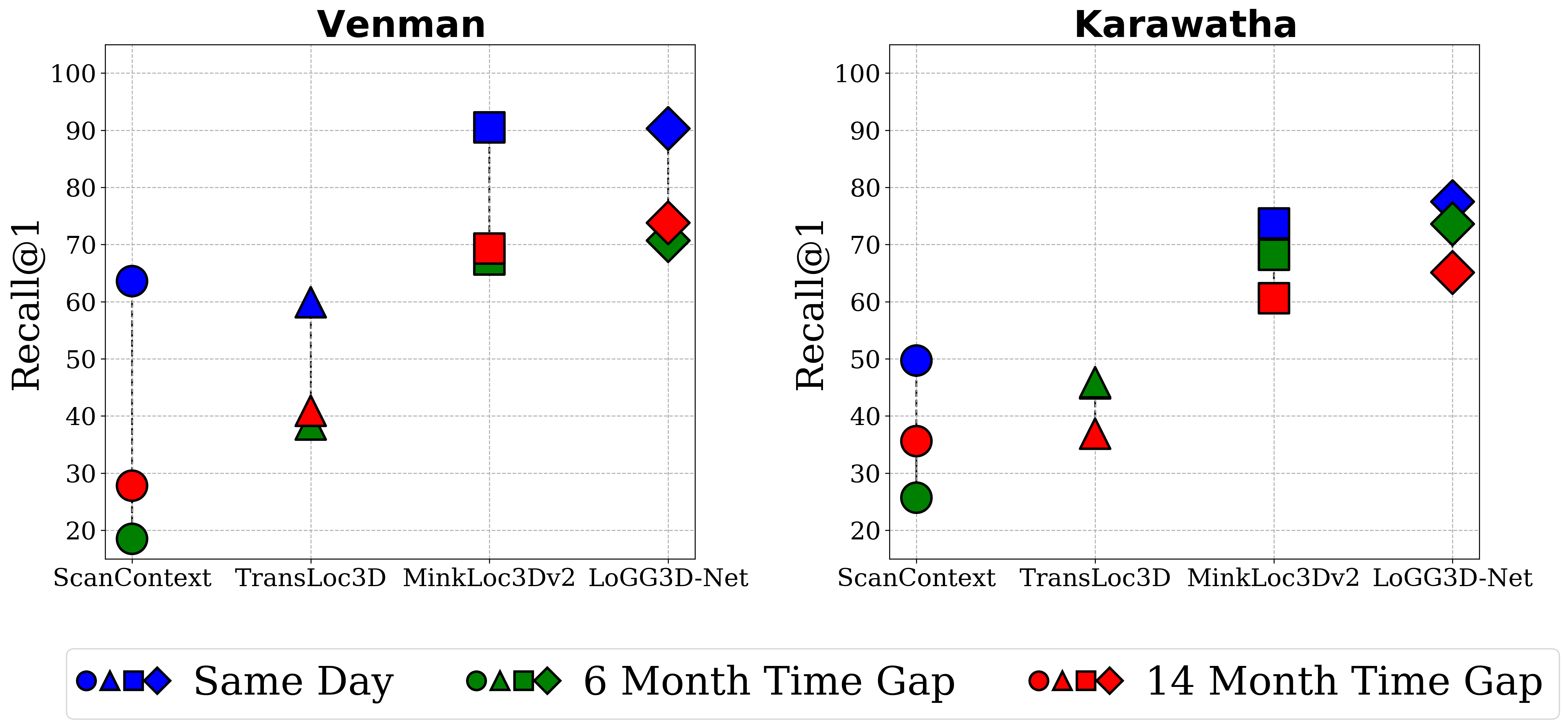}
    \caption{Impact of time elapsed between query and database sets on Recall@1 performance.  We use sequence-$01$ as the query and sequences $02$, $03$ and $04$ for the database sets to evaluate performance for time gaps of a same-day revisit, a 6-month revisit and a 14-month revisit respectively. }
    \label{fig:performance_drop}
    \vspace{-0.5cm}
\end{figure}

\textbf{Inter-Sequence Evaluation: } For inter-sequence evaluation we observe a similar drop in performance, with the strongest performing approach on Venman and Karawatha reporting only $79.84$\% and $74.67$\% top-1 Recall respectively, in contrast to the mid-to-high $90$\% reported in urban benchmarks.  Additionally, we investigate the impact that the time elapsed between the query and database sequences has on the performance of inter-sequence place recognition by using sequence $02$ as a query and evaluating retrieval performance against sequences $01$, $03$ and $04$ as databases representing a time gap of same day, 6 months and 14 months, respectively.  As shown in \figref{fig:performance_drop}, increasing the time gap between the query and the database has a significant detrimental impact on long-term place recognition performance, with top-1 Recall dropping by up to $45.1$\% and $24$\% in the Venman and Karawatha environments, respectively.  

\subsection{Discussion}
Natural environments not only contain highly irregular and unstructured features but more importantly change gradually over time, properties which are intuitively challenging for place recognition.  In our benchmarking we examine this hypothesis by testing multiple state-of-the-art approaches on our dataset, examining performance for intra-sequence and both short and long-term place recognition.  

For intra-sequence revisits we demonstrate that compared to performance on urban datasets, the state-of-the-art  approaches still have significant room for improvement on our dataset.  We also note that one of the most challenging aspects of intra-sequence evaluation in urban environments \--- reverse revisits \---  are significantly more difficult in dense forest environments.  This is seen in the large drop in performance on sequence V-03, which contains a large number of reverse revisits.  We hypothesise this is because in addition to the challenge posed by a changing viewpoint, in dense forests a larger portion of the environment has non-overlapping regions of occlusion during reverse revisits.

For inter-sequence evaluation we reveal that our dataset poses a significant challenge to state-of-the-art approaches, especially on the Karawatha environment where even the strongest performing approach only achieved a top-1 Recall performance in the mid $70$\%.  This challenge is especially prevalent for long-term inter-sequence evaluation, as top-1 recall is severely impacted when there is a significant time gap between the query and dataset sequences. 
We attribute this to the gradual changes in the environment over time, reducing the similarity between corresponding submaps as time progresses.

%% file: chapters/conclusion.tex
\section{Conclusion and Future Work}
In this paper, we release a new dataset for long-term place recognition known as the Wild-Places dataset, which is the first large-scale dataset for place recognition in unstructured, natural environments for both short and long term revisits.  We release roughly 63K point cloud submaps with accompanying 6DoF ground truth trajectories generated by our SLAM system, collected over the course of 14 months using a handheld sensor payload.  We provide an initial benchmarking of existing state-of-the-art approaches trained on our dataset for place recognition, and demonstrate the significant challenge posed by long-term revisits in dynamic unstructured environments.  For future work, we plan on extending the dataset to include semantic labels for point cloud segmentation to enable additional downstream tasks such as 3D scene understanding in natural environments.

%% file: main.bbl
\begin{thebibliography}{10}
\providecommand{\url}[1]{#1}
\csname url@rmstyle\endcsname
\providecommand{\newblock}{\relax}
\providecommand{\bibinfo}[2]{#2}
\providecommand\BIBentrySTDinterwordspacing{\spaceskip=0pt\relax}
\providecommand\BIBentryALTinterwordstretchfactor{4}
\providecommand\BIBentryALTinterwordspacing{\spaceskip=\fontdimen2\font plus
\BIBentryALTinterwordstretchfactor\fontdimen3\font minus
  \fontdimen4\font\relax}
\providecommand\BIBforeignlanguage[2]{{%
\expandafter\ifx\csname l@#1\endcsname\relax
\typeout{** WARNING: IEEEtran.bst: No hyphenation pattern has been}%
\typeout{** loaded for the language `#1'. Using the pattern for}%
\typeout{** the default language instead.}%
\else
\language=\csname l@#1\endcsname
\fi
#2}}

\bibitem{Geiger2013VisionMR}
A.~Geiger, P.~Lenz, C.~Stiller, and R.~Urtasun, ``{Vision meets robotics: The
  KITTI dataset},'' \emph{The International Journal of Robotics Research},
  vol.~32, pp. 1231 -- 1237, 2013.

\bibitem{kitti360}
Y.~Liao, J.~Xie, and A.~Geiger, ``{KITTI-360: A Novel Dataset and Benchmarks
  for Urban Scene Understanding in 2D and 3D},'' \emph{IEEE Transactions on
  Pattern Analysis and Machine Intelligence}, 2022.

\bibitem{carlevaris2016nclt}
N.~Carlevaris-Bianco, A.~K. Ushani, and R.~M. Eustice, ``{University of
  Michigan North Campus long-term vision and lidar dataset},'' \emph{The
  International Journal of Robotics Research}, vol.~35, no.~9, pp. 1023--1035,
  2016.

\bibitem{RobotCarDatasetIJRR}
W.~Maddern, G.~Pascoe, C.~Linegar, and P.~Newman, ``{1 Year, 1000km: The Oxford
  RobotCar Dataset},'' \emph{The International Journal of Robotics Research
  (IJRR)}, vol.~36, no.~1, pp. 3--15, 2017.

\bibitem{barnes2020oxford}
D.~Barnes, M.~Gadd, P.~Murcutt, P.~Newman, and I.~Posner, ``{The Oxford Radar
  RobotCar Dataset: A Radar Extension to the Oxford RobotCar Dataset},'' in
  \emph{2020 IEEE International Conference on Robotics and Automation
  (ICRA)}.\hskip 1em plus 0.5em minus 0.4em\relax IEEE, 2020, pp. 6433--6438.

\bibitem{jeong2019complexurban}
J.~Jeong, Y.~Cho, Y.-S. Shin, H.~Roh, and A.~Kim, ``Complex urban dataset with
  multi-level sensors from highly diverse urban environments,'' \emph{The
  International Journal of Robotics Research}, vol.~38, no.~6, pp. 642--657,
  2019.

\bibitem{Kim2020MulRanMR}
G.~Kim, Y.-S. Park, Y.~Cho, J.~Jeong, and A.~Kim, ``{MulRan: Multimodal Range
  Dataset for Urban Place Recognition},'' \emph{2020 IEEE International
  Conference on Robotics and Automation (ICRA)}, pp. 6246--6253, 2020.

\bibitem{lu2019l3}
W.~Lu, Y.~Zhou, G.~Wan, S.~Hou, and S.~Song, ``{L3-Net: Towards Learning Based
  LiDAR Localization for Autonomous Driving},'' in \emph{Proceedings of the
  IEEE/CVF Conference on Computer Vision and Pattern Recognition}, 2019, pp.
  6389--6398.

\bibitem{jiang2021rellis3d}
P.~Jiang, P.~Osteen, M.~Wigness, and S.~Saripalli, ``{RELLIS-3D Dataset: Data,
  Benchmarks and Analysis},'' in \emph{2021 IEEE international conference on
  robotics and automation (ICRA)}.\hskip 1em plus 0.5em minus 0.4em\relax IEEE,
  2021, pp. 1110--1116.

\bibitem{min2022orfd}
C.~Min, W.~Jiang, D.~Zhao, J.~Xu, L.~Xiao, Y.~Nie, and B.~Dai, ``{ORFD: A
  Dataset and Benchmark for Off-Road Freespace Detection},'' in \emph{2022
  International Conference on Robotics and Automation (ICRA)}.\hskip 1em plus
  0.5em minus 0.4em\relax IEEE, 2022, pp. 2532--2538.

\bibitem{Uy2018PointNetVLADDP}
M.~A. Uy and G.~H. Lee, ``{PointNetVLAD: Deep Point Cloud Based Retrieval for
  Large-Scale Place Recognition},'' \emph{2018 IEEE/CVF Conference on Computer
  Vision and Pattern Recognition}, pp. 4470--4479, 2018.

\bibitem{zhang2022hilti}
L.~Zhang, M.~Helmberger, L.~F.~T. Fu, D.~Wisth, M.~Camurri, D.~Scaramuzza, and
  M.~Fallon, ``{Hilti-Oxford Dataset: A Millimetre-Accurate Benchmark for
  Simultaneous Localization and Mapping},'' \emph{arXiv preprint
  arXiv:2208.09825}, 2022.

\bibitem{park2021elasticity}
C.~Park, P.~Moghadam, J.~L. Williams, S.~Kim, S.~Sridharan, and C.~Fookes,
  ``{Elasticity Meets Continuous-Time: Map-Centric Dense 3D LiDAR SLAM},''
  \emph{IEEE Transactions on Robotics}, vol.~38, no.~2, pp. 978--997, 2021.

\bibitem{tartandrive}
S.~Triest, M.~Sivaprakasam, S.~J. Wang, W.~Wang, A.~M. Johnson, and S.~Scherer,
  ``{TartanDrive: A Large-Scale Dataset for Learning Off-Road Dynamics
  Models},'' in \emph{2022 International Conference on Robotics and Automation
  (ICRA)}, 2022, pp. 2546--2552.

\bibitem{hudson2022heterogeneous}
N.~Hudson, F.~Talbot, M.~Cox, J.~Williams, T.~Hines, A.~Pitt, B.~Wood,
  D.~Frousheger, K.~Lo~Surdo, T.~Molnar, \emph{et~al.}, ``{Heterogeneous Ground
  and Air Platforms, Homogeneous Sensing: Team CSIRO Data61’s Approach to the
  DARPA Subterranean Challenge},'' \emph{Field Robotics}, vol.~2, no.~1, pp.
  595--636, 2022.

\bibitem{wigness2019rugd}
M.~Wigness, S.~Eum, J.~G. Rogers, D.~Han, and H.~Kwon, ``{A RUGD Dataset for
  Autonomous Navigation and Visual Perception in Unstructured Outdoor
  Environments},'' in \emph{2019 IEEE/RSJ International Conference on
  Intelligent Robots and Systems (IROS)}.\hskip 1em plus 0.5em minus
  0.4em\relax IEEE, 2019, pp. 5000--5007.

\bibitem{leyva2019tb}
M.~Leyva-Vallina, N.~Strisciuglio, M.~Lopez-Antequera, R.~Tylecek, M.~Blaich,
  and N.~Petkov, ``{TB-Places: A Data Set for Visual Place Recognition in
  Garden Environments.}'' \emph{IEEE Access}, vol.~7, no. 52277-52287, p.~2,
  2019.

\bibitem{valada2016deepmultispectraldataset}
A.~Valada, G.~L. Oliveira, T.~Brox, and W.~Burgard, ``{Deep Multispectral
  Semantic Scene Understanding of Forested Environments Using Multimodal
  Fusion},'' in \emph{International symposium on experimental robotics}.\hskip
  1em plus 0.5em minus 0.4em\relax Springer, 2016, pp. 465--477.

\bibitem{caesar2020nuscenes}
H.~Caesar, V.~Bankiti, A.~H. Lang, S.~Vora, V.~E. Liong, Q.~Xu, A.~Krishnan,
  Y.~Pan, G.~Baldan, and O.~Beijbom, ``{nuScenes: A Multimodal Dataset for
  Autonomous Driving},'' in \emph{Proceedings of the IEEE/CVF conference on
  computer vision and pattern recognition}, 2020, pp. 11\,621--11\,631.

\bibitem{chang2019argoverse}
M.-F. Chang, J.~Lambert, P.~Sangkloy, J.~Singh, S.~Bak, A.~Hartnett, D.~Wang,
  P.~Carr, S.~Lucey, D.~Ramanan, \emph{et~al.}, ``{Argoverse: 3D Tracking and
  Forecasting With Rich Maps},'' in \emph{Proceedings of the IEEE/CVF
  Conference on Computer Vision and Pattern Recognition}, 2019, pp. 8748--8757.

\bibitem{wilson2021argoverse}
B.~Wilson, W.~Qi, T.~Agarwal, J.~Lambert, J.~Singh, S.~Khandelwal, B.~Pan,
  R.~Kumar, A.~Hartnett, J.~K. Pontes, \emph{et~al.}, ``{Argoverse 2: Next
  Generation Datasets for Self-Driving Perception and Forecasting},'' in
  \emph{Thirty-fifth Conference on Neural Information Processing Systems
  Datasets and Benchmarks Track (Round 2)}, 2021.

\bibitem{steder2010frieburgdataset}
B.~Steder, G.~Grisetti, and W.~Burgard, ``Robust place recognition for 3d range
  data based on point features,'' in \emph{2010 IEEE International Conference
  on Robotics and Automation}.\hskip 1em plus 0.5em minus 0.4em\relax IEEE,
  2010, pp. 1400--1405.

\bibitem{sun2020scalabilitywaymo}
P.~Sun, H.~Kretzschmar, X.~Dotiwalla, A.~Chouard, V.~Patnaik, P.~Tsui, J.~Guo,
  Y.~Zhou, Y.~Chai, B.~Caine, \emph{et~al.}, ``{Scalability in Perception for
  Autonomous Driving: Waymo Open Dataset},'' in \emph{Proceedings of the
  IEEE/CVF conference on computer vision and pattern recognition}, 2020, pp.
  2446--2454.

\bibitem{pandey2011ford}
G.~Pandey, J.~R. McBride, and R.~M. Eustice, ``Ford campus vision and lidar
  data set,'' \emph{The International Journal of Robotics Research}, vol.~30,
  no.~13, pp. 1543--1552, 2011.

\bibitem{peynot2010marulan}
T.~Peynot, S.~Scheding, and S.~Terho, ``The marulan data sets: Multi-sensor
  perception in a natural environment with challenging conditions,'' \emph{The
  International Journal of Robotics Research}, vol.~29, no.~13, pp. 1602--1607,
  2010.

\bibitem{ramezani2020newer}
M.~Ramezani, Y.~Wang, M.~Camurri, D.~Wisth, M.~Mattamala, and M.~Fallon, ``{The
  Newer College Dataset: Handheld LiDAR, Inertial and Vision with Ground
  Truth},'' in \emph{2020 IEEE/RSJ International Conference on Intelligent
  Robots and Systems (IROS)}.\hskip 1em plus 0.5em minus 0.4em\relax IEEE,
  2020, pp. 4353--4360.

\bibitem{vidanapathirana2021locus}
K.~Vidanapathirana, P.~Moghadam, B.~Harwood, M.~Zhao, S.~Sridharan, and
  C.~Fookes, ``{Locus: LiDAR-based Place Recognition using Spatiotemporal
  Higher-Order Pooling},'' in \emph{2021 IEEE International Conference on
  Robotics and Automation (ICRA)}.\hskip 1em plus 0.5em minus 0.4em\relax IEEE,
  2021, pp. 5075--5081.

\bibitem{komorowski2021egonn}
J.~Komorowski, M.~Wysoczanska, and T.~Trzcinski, ``{EgoNN: Egocentric Neural
  Network for Point Cloud Based 6DoF Relocalization at the City Scale},''
  \emph{IEEE Robotics and Automation Letters}, vol.~7, no.~2, pp. 722--729,
  2021.

\bibitem{vidanapathirana2021logg3d}
K.~Vidanapathirana, M.~Ramezani, P.~Moghadam, S.~Sridharan, and C.~Fookes,
  ``{LoGG3D-Net: Locally Guided Global Descriptor Learning for 3D Place
  Recognition},'' \emph{International Conference on Robotics and Automation},
  2022.

\bibitem{cattaneo2022lcdnet}
D.~Cattaneo, M.~Vaghi, and A.~Valada, ``{LCDNet: Deep Loop Closure Detection
  for LiDAR SLAM based on Unbalanced Optimal Transport},'' \emph{IEEE
  Transactions on Robotics}, 2022.

\bibitem{komorowski2022improving}
J.~Komorowski, ``Improving point cloud based place recognition with
  ranking-based loss and large batch training,'' \emph{2022 International
  Conference on Patern Recognition (ICPR)}, 2022.

\bibitem{Vidanapathirana2023SpectralGV}
K.~Vidanapathirana, P.~Moghadam, S.~Sridharan, and C.~Fookes, ``{Spectral
  Geometric Verification: Re-Ranking Point Cloud Retrieval for Metric
  Localization},'' \emph{IEEE Robotics and Automation Letters}, 2023.

\bibitem{shaban2022semantic}
A.~Shaban, X.~Meng, J.~Lee, B.~Boots, and D.~Fox, ``{Semantic Terrain
  Classification for Off-Road Autonomous Driving},'' in \emph{Conference on
  Robot Learning}.\hskip 1em plus 0.5em minus 0.4em\relax PMLR, 2022, pp.
  619--629.

\bibitem{baril2021kilometer}
D.~Baril, S.-P. Desch{\^e}nes, O.~Gamache, M.~Vaidis, D.~LaRocque, J.~Laconte,
  V.~Kubelka, P.~Gigu{\`e}re, and F.~Pomerleau, ``Kilometer-scale autonomous
  navigation in subarctic forests: challenges and lessons learned,''
  \emph{arXiv preprint arXiv:2111.13981}, 2021.

\bibitem{ramezani2022wildcat}
M.~Ramezani, K.~Khosoussi, G.~Catt, P.~Moghadam, J.~Williams, P.~Borges,
  F.~Pauling, and N.~Kottege, ``{Wildcat: Online Continuous-Time 3D
  Lidar-Inertial SLAM},'' \emph{arXiv preprint arXiv:2205.12595}, 2022.

\bibitem{torr2000mlesac}
P.~H. Torr and A.~Zisserman, ``Mlesac: A new robust estimator with application
  to estimating image geometry,'' \emph{Computer vision and image
  understanding}, vol.~78, no.~1, pp. 138--156, 2000.

\bibitem{kim2018scan}
G.~Kim and A.~Kim, ``{Scan Context: Egocentric Spatial Descriptor for Place
  Recognition Within 3D Point Cloud Map},'' in \emph{2018 IEEE/RSJ
  International Conference on Intelligent Robots and Systems (IROS)}, 2018, pp.
  4802--4809.

\bibitem{Xu2021TransLoc3DP}
T.-X. Xu, Y.-C. Guo, Y.-K. Lai, and S.-H. Zhang, ``{TransLoc3D : Point Cloud
  based Large-scale Place Recognition using Adaptive Receptive Fields},'' in
  \emph{International Conference on Computational Visual Media}, 2022.

\end{thebibliography}
